\documentclass[journal]{IEEEtran}
\usepackage{subfigure}
\usepackage{graphicx}
\usepackage{setspace}
\usepackage{amsmath}
\usepackage{amsfonts}
\usepackage{float}

\usepackage[ruled,longend]{algorithm2e}
\usepackage{geometry} 

\begin{document}

\title{RandCrowns: A Quantitative Metric for Imprecisely Labeled Tree Crown Delineation}

\author{Dylan~Stewart,~\IEEEmembership{Member,~IEEE,}
       Alina~Zare,~\IEEEmembership{Senior~Member,~IEEE,}
       Sergio~Marconi, Ben~G.~Weinstein, Ethan~P.~White, Sarah~J.~Graves, Stephanie~A.~Bohlman, Aditya~Singh \thanks{Dylan Stewart and Alina Zare are with the Department of Electrical and Computer Engineering, University of Florida, Gainesville, FL 32611 USA (email: azare@ece.ufl.edu).} \thanks{Ben Weinstein and Ethan White are with the Department of Wildlife Ecology and Conservation, University of Florida, Gainesville, FL 32611 USA.} \thanks{Sarah Graves is with the Nelson Institute for Environmental Studies, University of Wisconsin-Madison, Madison, WI 53706 USA.} \thanks{Sergio Marconi and Stephanie Bohlman are with the School of Forest, Fisheries, and Geomatics Sciences, University of Florida, Gainesville, FL 32611 USA.} \thanks{Aditya Singh is with the Department of Agricultural and Biological Engineering, University of Florida, Gainesville, FL 32611 USA.}}


\maketitle

\begin{abstract}
Supervised methods for object delineation in remote sensing require labeled ground-truth data. Gathering sufficient high quality ground-truth data is difficult, especially when targets are of irregular shape or difficult to distinguish from background or neighboring objects. Tree crown delineation provides key information from remote sensing images for forestry, ecology, and management. However, tree crowns in remote sensing imagery are often difficult to label and annotate due to irregular shape, overlapping canopies, shadowing, and indistinct edges. There are also multiple approaches to annotation in this field (e.g., rectangular boxes vs. convex polygons) that further contribute to annotation imprecision. However, current evaluation methods do not account for this uncertainty in annotations, and quantitative metrics for evaluation can vary across multiple annotators. In this paper, we address these limitations by developing an adaptation of the Rand index for weakly-labeled crown delineation that we call RandCrowns. Our new RandCrowns evaluation metric provides a method to appropriately evaluate delineated tree crowns while taking into account imprecision in the ground-truth delineations. The RandCrowns metric reformulates the Rand index by adjusting the areas over which each term of the index is computed to account for uncertain and imprecise object delineation labels. Quantitative comparisons to the commonly used intersection over union method shows a decrease in the variance generated by differences among multiple annotators. Combined with qualitative examples, our results suggest that the RandCrowns metric is more robust for scoring target delineations in the presence of uncertainty and imprecision in annotations that are inherent to tree crown delineation.    
\end{abstract}

\begin{IEEEkeywords}
Remote sensing, Tree crown delineation, Imprecise labels, Quantitative evaluation
\end{IEEEkeywords}

\IEEEpeerreviewmaketitle
\section{Introduction}\label{Introduction}
\IEEEPARstart{I}{dentifying} trees in remote sensing imagery is essential to understanding individual-level ecological processes at large scales. Delineating individual tree crowns is required for research on topics ranging from plant traits \cite{Marconi2021}, to carbon sequestration \cite{Dalponte2016,fao,Graves2018}, to biodiversity assessment \cite{Graves2016,Jaafar2018,Saarinen2018}. As a result, an array of methods have been developed for converting a variety of remote sensing data into information on individual trees. These include LiDAR-based methods that use three-dimensional point cloud representations of forest structure to cluster points into individual trees. These LiDAR-based methods include counting trees within segments by finding the number of unique Gaussian distributions needed to recreate the LiDAR data with maximum correlation \cite{Hu14}, splitting segments which harbor multiple trees into individual delineations by finding saddle points in 2D contour maps  \cite{Wu2016}, and merging watershed segments to delineate individual trees using a minimum spanning forest \cite{Mongus2015}. There are also methods based on RGB imagery that identify groups of pixels that match the appearance of a tree from above, including a convolutional neural network pretrained on tens of millions of LiDAR-annotated crowns and fine-tuned using hand-labelled crowns in the RGB imagery \cite{Weinstein2019} and a multi-sensor fusion approach of RGB images with LiDAR intensity and hand-crafted features \cite{Deng2016}. Given the large number of methods currently available for delineating tree crowns, a key question is how best to evaluate the performance of different approaches.

Evaluation of crown delineation algorithms is currently accomplished using a variety of methods including: 1) comparisons of how many trees are delineated in a plot to the number counted in the field \cite{Wang04}; 2) assessment of whether individual delineated crowns are associated with a field mapped stem \cite{Lucas2006}; and 3) direct comparison of how similar delineated crowns are to expert labeled crowns \cite{Weinstein2019}. Human labeled crowns are generated either by visually identifying crowns on the remote sensing imagery or by performing this task while in the field observing the trees being labeled \cite{Graves2018b}. Direct comparisons of algorithmic and human delineated crowns provides the most challenging assessment of the underlying methods. Comparisons between labeled and delineated crowns are often made based on the Intersection over Union (IoU) score, also known as the Jaccard index \cite{Jaccard1908}. The IoU is the ratio of the area encompassed by the intersection of two sets divided by the union of the pair. Another related method to assess agreement between multiple annotations is the Rand index \cite{Rand1971}. The Rand index relates the ratio of pairs of points that are grouped into the same segment (or delineation) to the total number of labeled pairs. These fundamental measures can also be converted to recall (and precision if all crowns are delineated) using an IoU/Rand index threshold.

While IoU is a common measure of overlap or similarity when delineating objects, it assumes that the ground-truth is a precise representation of the true crown. As a result, it does not account for the potentially large measurement error in human-labeled crown boundaries. Boundaries of tree crowns are difficult for humans to delineate due to imprecise boundaries, complex three dimensional structure and object occlusion. Especially in dense forests, it can be difficult to determine whether branches belong to a target crown or neighboring crown.  When relying solely on remote sensing imagery, it can also be difficult to tell the difference between single large trees and multiple smaller trees because of how small trees may cluster together, or how large trees may have multiple clusters of branches \cite{Graves2018}. This is likely to result in significant variation in ground-truth labels depending on the person doing the labeling and the labeling method. Some of these issues can be mitigated by labeling the imagery while observing the tree in the field \cite{Graves2018b}. However, there is likely significant variation among observers in the specific crowns delimited on the scale of a few meters. Additionally, field based labeling is time consuming, making labeling based only on imagery alone an important tool. This measurement error (or “weak labeling”) influences the IoU score and current methods do not address this challenge.

The issue with evaluating weakly labeled segmentations is a common challenge in remote sensing and image segmentation more broadly \cite{Stewart2021}. There are a variety of approaches to addressing measurement error for semantic segmentation. Global and Local Consistency Error (GCE and LCE respectively) compute changes in segmentation membership across multiple partitionings \cite{Martin2001}. The only cases where GCE and LCE award zero error are: each segment in an image consists of a single pixel or the entire image contains one segment. Variation of Information (VOI) measures the mutual information between two segmentations \cite{Meila2005}. VOI is most appropriate when segment membership probabilities are assigned to each pixel. The Probabilistic Rand Index alters the Rand index by scaling correct pairs by their average number of occurrences within multiple ground-truth labelings \cite{Unnikrishnan2005}. While these methods attempt to address variance across multiple paritionings of semantic labels, none of these approaches directly address scoring in relation to imprecise and weakly-labeled ground-truth. 

In this paper we develop a new metric, RandCrowns (RC), for evaluating overlap between algorithmic and human delineated tree crowns that takes into account the specific kinds of uncertainty and imprecision associated with tree crown delineation. Due to the imprecision and uncertainty of the desired delineations, we refer to the desired tree crown not as ground-truth, but as a desired target. This approach builds on the Rand index by using “boundaries”, or buffers, around the desired crown that provide a cushion to accommodate variation in the labels. The  goal  of  the  RandCrowns  measure  is  to  appropriately  score  delineations  which encompass  a  desired  crown  regardless  of  realistic differences  in  labels that do not meaningfully change the labels provided. It is important to note that a relationship exists between appropriately scoring a variety of candidate delineations and assigning large scores to those with questionably large differences. This is a balance that can be controlled in our measure by modifying parameter settings. Other current methods do not allow for any balance of this kind. Although our method does require suitable selection of parameters, it is a tradeoff for an overall improvement in reliability. We evaluate this measure and compare it to the most commonly used metric for individual crown level comparisons, the IoU, using a dataset of tree crowns annotated by multiple human labelers. We find that when properly parameterized, the RandCrowns measure results in reduced variance in score resulting from differences among labelers and therefore provides a more robust approach to assessing crown delineation algorithms at the fundamental scale of the individual tree. While we develop and apply the RandCrowns measure to tree crown delineation, it can be used to analyze performance for any segmentation application whereby labels contain uncertainty (as is the case in many remote sensing problems).

\section{RandCrowns}\label{sec:RandCrowns}
\begin{figure*}[h!]
\subfigure[]{\includegraphics[width=0.45\linewidth]{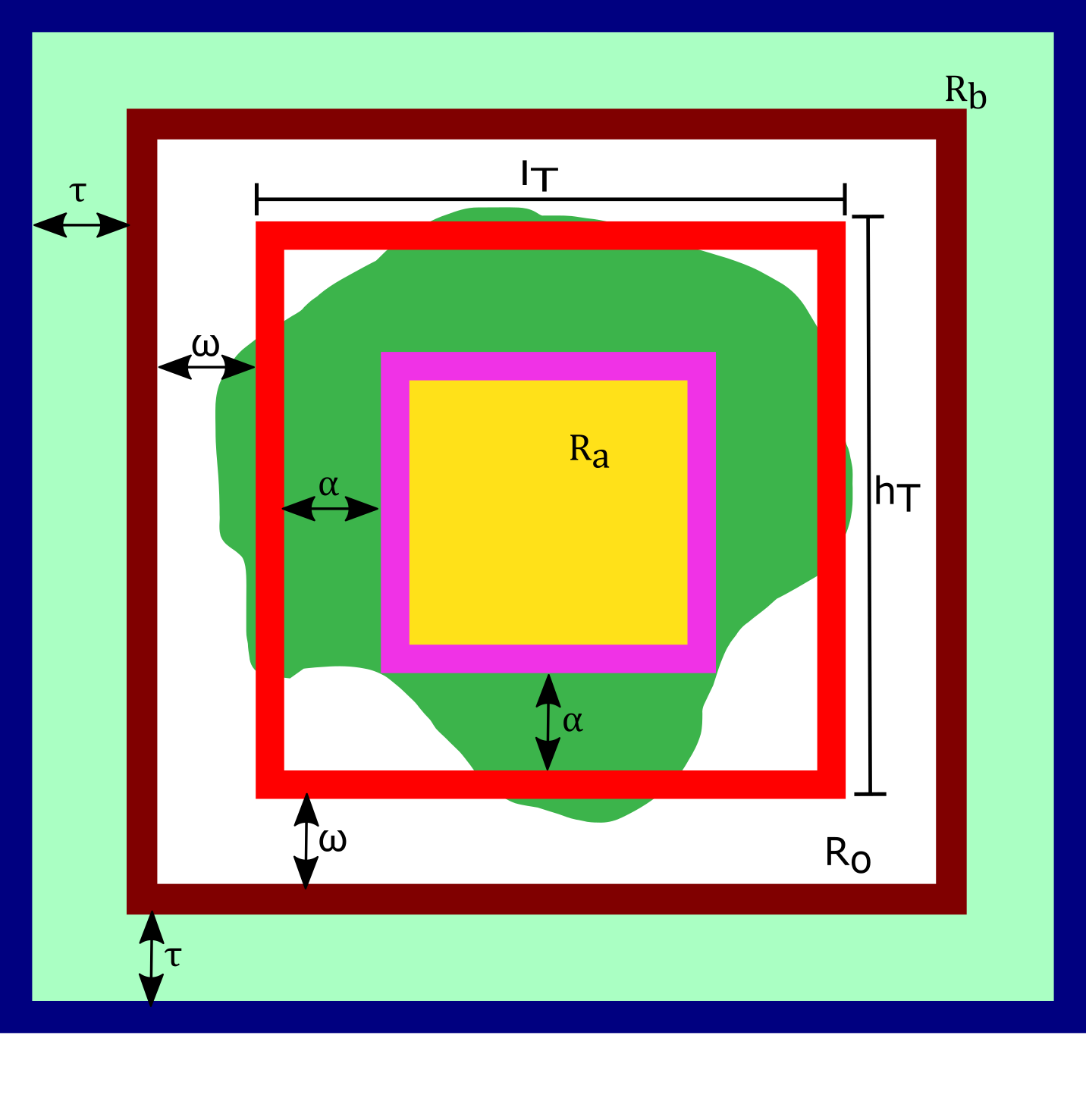} \label{fig:RCa}} \hfill
\subfigure[]{\label{fig:RCb} \includegraphics[width=0.45\linewidth]{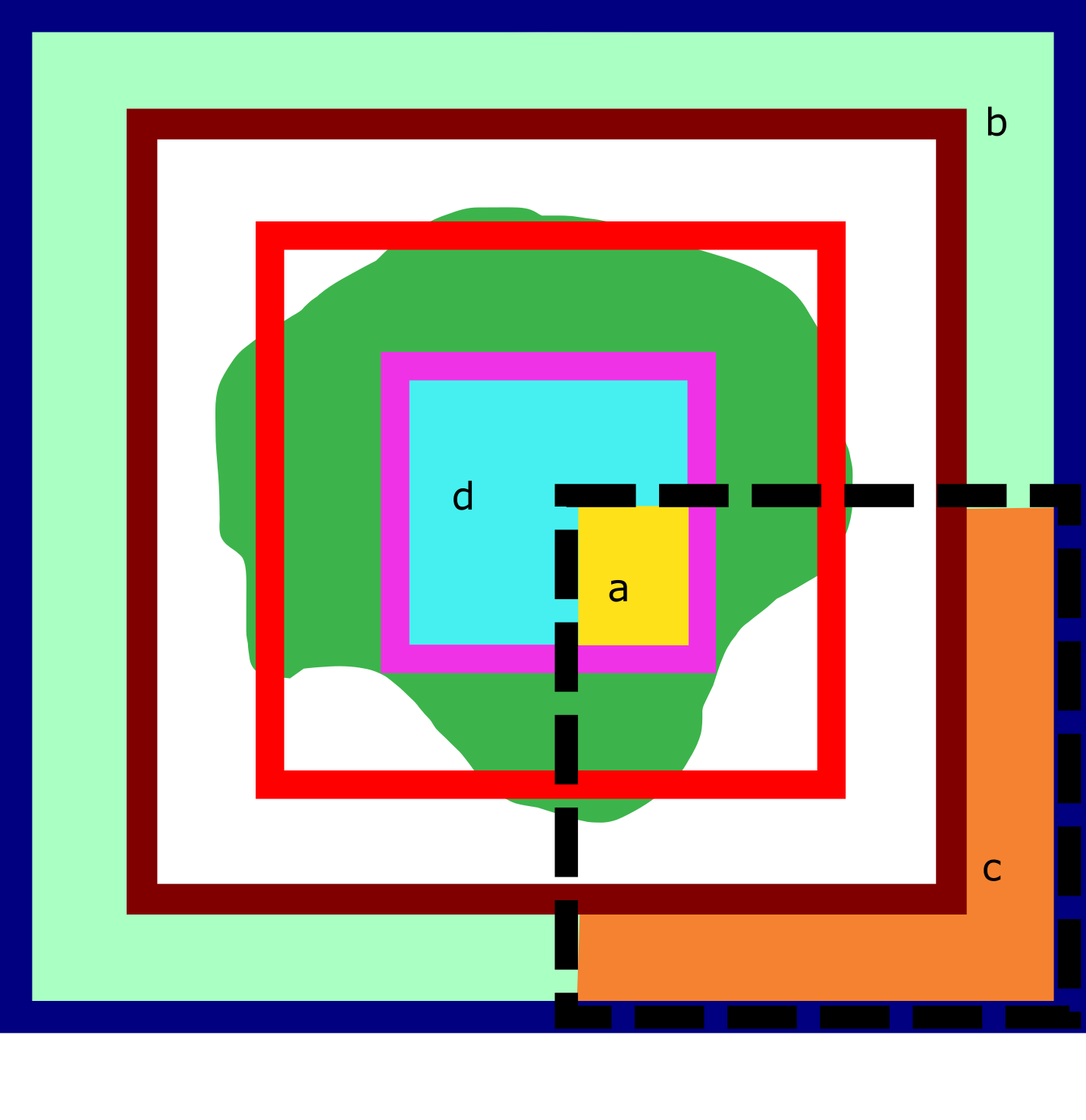}} \\
\begin{center}
\subfigure{\includegraphics[width=.75\linewidth]{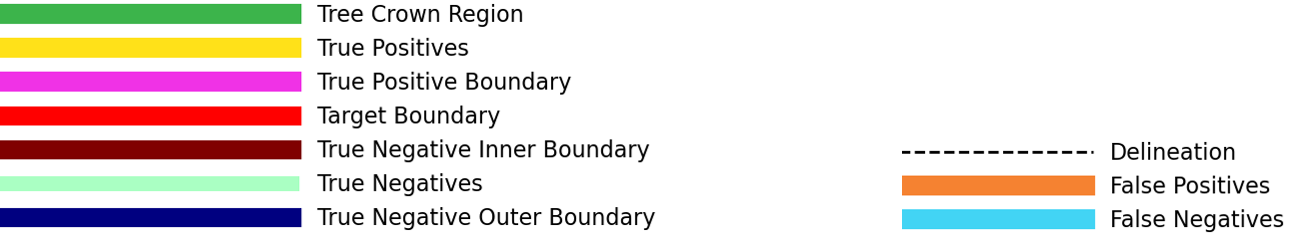}}
\end{center}
\caption{A diagram of the RandCrowns regions and corresponding boundaries. An example crown is shown in green. In Figure 1a, the desired target from an annotator is shown in red and the length and height are denoted $l_T$ and $h_T$ respectively. The true positive boundary is within the desired target by $2\alpha$ pixels in both length and height. The yellow region within the true positive boundary represents the true positive region (denoted “a” in the Rand equation) and is labeled  $R_a$. The true negative inner boundary is shown in maroon and is larger by $2\omega$ in both length and height than the desired target. A region that is ignored in the computation of the RandCrowns measure is set based on the true negative inner boundary, $R_o$. Outside of the true negative inner boundary is the true negative outer boundary shown in navy. The true negative outer boundary envelops the true negative region, $R_b$  (denoted as “b” in the Rand equation), to be $2\tau$ larger than the true negative inner boundary in both length and height.  It is important to note within the red delineation there are some gaps where there is no crown, this represents the imprecise labeling for an imperfect “true” crown delineation. In Figure 1b, the black box is a sample delineation. The true positives are shown in the yellow highlighted region (labelled “a”); this is the intersection of the delineation and the true positive boundary. The true negatives are the mint highlighted region (labelled “b”). True negatives are the difference between the true negative outer boundary and the true negative inner boundary and the delineation. The false positives are in the orange highlighted region (labelled “c”). This region shows the portion of the delineation that is contained in $R_b$. The false negatives are in the cyan region (labelled “d”). This region shows the difference between the true positive boundary and the delineation.}
\label{fig:RandCrowns}
\end{figure*}

The traditional Rand index (RI) is a measure of similarity between two partionings of data. The RI compares the ratio of how many pairs of points in two partitionings agree to how many pairs of points disagree,

\begin{equation}
    \label{eq:Rand}
     \textrm{Rand} = \frac{a^\prime+b^\prime}{a^\prime+b^\prime+c^\prime+d^\prime},
\end{equation}
where $a^\prime$ is the number of pairs of points that are found in the same segment in both partitions and $b^\prime$ is the number of pairs of points that are found in different segments in both partitions. The terms $c^\prime$ and $d^\prime$ are the number of pairs of points that are together in one partitioning but not in the other and vice versa. This is similar to the IoU metric which is often used to measure the overlap between predicted and observed tree crowns. However, IoU does not consider a term analogous to $b^\prime$, nor does the IoU account for any uncertainty. IoU can be written with the terms from the RI as, 

\begin{equation}
    \label{eq:IoU}
    \textrm{IoU} = \frac{a^\prime}{a^\prime+c^\prime+d^\prime}.
\end{equation}

The original RI, like the IoU, is simply a comparison measure between any pair of partitionings (delineations). When evaluating crown delineations, this overlap between the predicted delineation and the ground-truth represents how accurately the delineation performed. As discussed above, ground-truth crowns are typically delineated either in the field (while looking at remote sensing imagery) or by labeling remote sensing imagery alone (without field observation). If these ground-truth crowns were delineated without error then either the RI or the IoU would provide consistent measures of the individual level accuracy of the crown delineation. However, the difficulties in identifying the edges of tree crowns result in imprecision in the ground-truth labels and significant variation between human annotators when labeling the same tree crowns. Therefore, we assume that there is uncertainty to each delineated crown. 

For each tree, the region drawn by a human annotator that encompasses the area of a tree may be larger or smaller than the actual canopy. This region, known as the desired target, is defined as in Equation \ref{eq:Target},

\begin{equation}
    \label{eq:Target}
    \begin{split}
    T = \Big\{(x,y) \in \mathbb{R}^2 : &|x-x_T|\leq \frac{l_T}{2},\\ 
    &|y-y_T|\leq \frac{h_T}{2}\Big\},
    \end{split}
\end{equation}

\begin{table*}[ht!]
    \begin{equation}
    \label{eq:Delineation}
   D =\Big\{(x,y)\in \mathbb{R}^2 : |x - x_D| \leq \frac{l_D}{2},|y - y_D| \leq \frac{h_D}{2}\Big\}
    \end{equation}
    
    \begin{equation}
    \label{eq:TPR}
    R_{a} =\Big\{(x,y) \in \mathbb{R}^2 : |x-x_T|\leq \frac{l_T}{2}-\alpha, |y-y_T|\leq \frac{h_T}{2}-\alpha\Big\}
\end{equation}

\begin{equation}
    \label{eq:OR}
        R_{o} = \Big\{(x,y) \in \mathbb{R}^2 : \frac{l_T}{2}< |x-x_T|\leq \frac{l_T}{2}+\omega, \frac{h_T}{2}<|y-y_T|\leq \frac{h_T}{2}+\omega\Big\}
\end{equation}

\begin{equation}
    \label{eq:OuterBoundaryParameter}
    |R_a| = (h_T+2\omega+2\tau)(l_T+2\omega+2\tau)-h_Tl_T
\end{equation}

 \begin{equation}
    \label{eq:EdgeRegion}
    R_{e} = \Big\{(x,y) \in \mathbb{R}^2 : \frac{l_T}{2}+\omega<|x-x_T|\leq \frac{l_T}{2}+\omega+\gamma \tau,
    \frac{h_T}{2}+\omega <|y-y_T|\leq\frac{h_T}{2}+\omega+\gamma \tau\Big\}
    \end{equation}
\end{table*}

where $(x_T,y_T)$ are the center coordinates of the desired target and $l_T$ and $h_T$ are the length and height respectively. To deal with the imprecision of this labeling, in the RandCrowns measure, the true positive region ($R_a$, analogous to $a^\prime$) is concentric with the desired target and spans a smaller, less uncertain area within the annotated region. Due to the possibility of delineations being larger than the actual tree, the true negative region ($R_b$, analogous to $b^\prime$) is concentric with the desired target but spans an area that is buffered outside of the desired boundary. 

Consider a delineation, $D$, where the center of the delineation is indicated by $(x_D,y_D)$ with length and height being $l_D$ and $h_D$, respectively as shown in Equation \ref{eq:Delineation}. To quantify the relationship between the delineated and desired target, we modify each term of the Rand index to be applied to the true positive and true negative regions and denote this adaptation as the RandCrowns score. The regions for the RandCrowns index are shown in Figure \ref{fig:RandCrowns}.

For a given desired target, the true positive region, $R_{a}$, is set based on a parameter, $\alpha$, as shown in Equation \ref{eq:TPR}, with $\alpha>0$. The true positive region is within the desired target and is a refined area of the tree crown for which there is more certainty that the crown is encompassed there. Outside of the desired target boundary there is an outer region.

The outer region represents the area of uncertainty that the desired target could extend to, but with high certainty does not extend outside of this area. The outer region, $R_{o}$, is computed based on the parameter, $\omega$, and the desired target as shown in Equation \ref{eq:OR}, with $\omega>0$. The outer region extends outside of the desired target and is used to produce the true negative region.

To properly penalize delineations that are much larger than the tree crown, the area of the true negative region is set to be a multiple, $\gamma$, of the true positive region area. Given the area of the true positive region and the length and height of the outer region, a direct solution to produce the true negative outer boundary parameter relies on the quadratic equation in Equation \ref{eq:OuterBoundaryParameter}, where $|R_a|$  denotes the area of the true positive region. The difference between the edge length and height and that of the outer region ($\tau$ in Figure \ref{fig:RandCrowns}) can be used to set the area of the true negative region, $R_b$. Solving for $\tau$ and adding multiples of $\tau$ to the length and height of the outer region produces an intermediate edge region in Equation \ref{eq:EdgeRegion}.

In order to appropriately score large delineations that contain portions of the inner region and extend outside the true negative outer boundary, we extend the edge region for delineations that satisfy the condition: $R_e^* =  R_e \cup D$.  If the delineation does not extend outside of the true negative region, the region is not modified. The true negative region is computed based on the edge and outer regions: $R_b = R_e^* \setminus R_o$. Based on the solution for the edge region, changing $\gamma$ in Equation \ref{eq:EdgeRegion} produces an area ratio in Equation \ref{eq:AreaRatio},

\begin{equation}
    \label{eq:AreaRatio}
     \frac{|R_b|}{|R_a|}=\gamma.
\end{equation}

For irregularly shaped desired targets, the true negative outer boundary can be set heuristically through relating the areas of the true positive and resultant true negative region. The true negative region is set by buffering the outer region iteratively and checking the area ratio of the produced true negative region to the true positive region until the ratio satisfies the desired value $\gamma$. When the predefined $\gamma$ value would expand the true negative outer boundary outside of the image area, these unavailable pixels are clipped and only those existing in the image area are used. The steps to produce the edge region, and subsequently the true negative region that satisfies the area ratio for irregularly shaped desired targets is shown in Algorithm \ref{alg:PseudoCodeRC}. 

\begin{algorithm}
\label{alg:PseudoCodeRC}
  $\text{Initialize } \epsilon = 2\gamma\omega$\;
  $R_e \gets R_o.\text{buffer}(\epsilon)$\;
  \While{$\frac{|R_b|}{|R_a|}>\gamma$}
  {
    $\epsilon \gets \epsilon - \delta$\;
    $R_e \gets R_o.\text{buffer}(\epsilon)$\;
  }
  \KwRet{$R_e$}\;
  \caption{Compute $R_e$ to satisfy $\frac{|R_b|}{|R_a|}=\gamma$}
\end{algorithm}

The RandCrowns measure is developed by modifying each of the four terms of the RI to be appropriate for imprecisely labeled targets. The correct number of true positive pairs between a detected delineation and a desired labeled target is computed by the intersection of the delineation and the true positive region as shown in Equation \ref{eq:RandCrownsAterm},

\begin{equation}
    \label{eq:RandCrownsAterm}
    a = |D \cap R_a|^2.
\end{equation}

This term will promote delineations that completely enclose the true positive region of the desired target. If a delineation completely includes the true positive region then this true positive score will be maximized. Whereas if a delineation does not consist of the entire portion of the true positive region, the true positive score will be smaller.

RandCrowns ensures that not only do detected delineations encapsulate the true positive region, but also that delineations do not extend past the target encompassing background non-target regions. This is evaluated by computing the number of correct true negative pairs. The number of correct true negative pairs are computed as shown in Equation \ref{eq:RandCrownsBterm},

\begin{equation}
    \label{eq:RandCrownsBterm}
    b = |R_b \setminus D|^2.
\end{equation}

If the delineation does not surround portions of the true negative region, $b$ will be high for a given delineation. If the delineation expands outside of the true negative inner boundary, $b$ will decrease. Thus, $b$ promotes delineations that do not extend outside the imprecise desired target. The false positive pairs and false negative pairs can also be computed.

To account for oversegmentations where the delineation contains regions outside of the desired crown area, the false positive pairs are counted. The false positive pairs are counted from the pixels within a delineation that cross into the true negative region as shown in Equation \ref{eq:RandCrownsCterm},

\begin{equation}
    \label{eq:RandCrownsCterm}
    c = |D \cap R_b|^2.
\end{equation}

If a delineation envelops portions outside of the true negative inner boundary then the false positive score will be high. Conversely, if the delineation only encloses portions within the true negative inner boundary, $c$ will be low. This promotes small bounding boxes that do not include the true negative region surrounding the imprecisely labeled desired target.

Lastly, the number of false negative pairs are counted. These pixels are portions of the desired crown area that are missing from the delineation. The number of false negative pairs are pixels within the true positive region that are not within the delineation as shown in Equation \ref{eq:RandCrownsDterm},

\begin{equation}
    \label{eq:RandCrownsDterm}
    d = |R_a \setminus D|^2.
\end{equation}

The number of false negative pairs will be small if the delineation encompasses most of the true positive region. When smaller areas of the true positive region are enclosed within the delineation, the false negative score will be larger. Analogous to the Rand index, we combine the four terms defined above to create the RandCrowns index in Equation \ref{eq:RandCrownsFull},

\begin{equation}
    \label{eq:RandCrownsFull}
    \textrm{RandCrowns} = \frac{a+b}{a+b+c+d},
\end{equation}

where the terms $a,b,c,\textrm{and }d$ are defined for imprecise desired targets.

\section{Data}\label{sec:Data}
For this project we used orthorectified RGB photos collected from aircraft (0.1 m spatial resolution). The data was collected from the National Ecological Observatory Network Airborne Observatory Platform (NEON AOP) and clipped in four 40$\times$40 m plots. Data was collected from three different forest ecosystems to better estimate the effect of forest type on crown annotation uncertainty and imprecision. Plots in the Ordway-Swisher Biological Station (OSBS) represent an open forest (with tree crowns relatively sparse); plots in the Mountain Lake Biological Station (MLBS) represent a generally closed forest (with crowns generally overlapping and more difficult to visually separate); plots in Talladega National Forest (TALL) represent a mix of open and closed forests. Two sets of annotations were used: one consists of polygons derived from field-measurements while the other was produced from remote sensing delineations.

The field polygons were generated by marking the location and boundary of individual tree crowns in NEON aerial photos while in the field \cite{Graves2018}. Individual trees were located in the field plots using a GPS and the images for reference. The crown boundaries were digitized on the images to create the tree crown objects. Delineated crowns were refined in the lab using multiple NEON data products. These field delineations are available as part of a benchmark dataset for tree crown delineation \cite{Weinstein2020}. The second set of annotations were labeled by four different experts using images alone, not in the field. Each was asked to hand delineate all visible crowns directly on each plot image. QGIS software was used to generate rectangular bounding boxes using a combination of RGB, hyperspectral and LiDAR data, following the approach proposed by \cite{Weinstein2020}. Annotators were not allowed to view the annotations collected from the other three experts, to ensure independence and better estimate uncertainty in crown delineation among multiple annotators. 

Due to the lack of a single true ground-truth due to variation among multiple annotations, we cross-validate delineations from four annotators where each annotator is considered the desired target and the other three are sample delineations. The delineations are from four diverse plots containing multiple trees per plot.  Each plot is 40$\times$40 m with 10 cm resolution from the NEON dataset. The total number of annotations for each annotator is listed in Table \ref{table:Dataset}.

\begin{table}[t]
\centering
\caption{Experiment 1 dataset. Four plots from various geographic locations were labeled by four annotators.}
\label{table:Dataset}
\begin{tabular}{|l|l|l|l|l|l|}
\hline
Plot        & 1  & 2  & 3  & 4  & Total \\ \hline
Annotator 1 & 48 & 43 & 26 & 35 & 152   \\ \hline
Annotator 2 & 48 & 54 & 31 & 58 & 191   \\ \hline
Annotator 3 & 64 & 56 & 60 & 63 & 243   \\ \hline
Annotator 4 & 70 & 33 & 34 & 54 & 191   \\ \hline
\end{tabular}
\end{table}

The annotations were reduced to remove clearly mismatched annotations between the labelers (i.e., cases where the labels were for clearly different tree crowns). All annotations from each plot were visualized simultaneously by overlaying them on the RGB raster using QGIS and delineations of which there was little or no qualitative agreement among multiple annotators were removed. Within the raw data, multiple cases of unshared labels are
present. In some instances, a subset of labelers did not label crowns that other annotators
did. In other cases, one annotator labeled what visually appeared to be a single large
crown with four separate canopies. However, other annotators labeled each of the four
separate canopies as four individual crowns. Additionally, a pair of annotators did not
label some small crowns that the other pair did, so these unshared delineations were also
removed. In any case, where not all four annotators provided labels for the same crowns,
these annotations were removed. We removed these annotations in order to rely on the
performance of all possible annotators to assist for parameter selection. Without
removing unshared delineations, each expert would not have the same weight in
parameter selection. The mean crown size of Experiment 1 delineations was 39.6 m$^2$ and standard deviation 22.0 m$^2$.  A similar dataset was prepared given field-delineated annotations and two sets of annotations derived from the remote sensing data. The mean crown size of Experiment 2 field-delineations was 22.3 m$^2$ and standard deviation 15.8 m$^2$.

\section{Experiments}\label{sec:Experiments}
To demonstrate our new RandCrowns method we provide two unique experiments. The first is a cross-validation experiment which compares multiple annotations of the same tree crowns delineated via remote sensing imagery by four annotators. Parameters for the model are modified to reduce the variance between annotators. While small variance between annotators is desired, qualitative results are considered influential to the selection of RandCrowns parameters. The second experiment applies the RandCrowns measure to a dataset consisting of field delineated crowns and scores bounding box annotations derived from the remote sensing data. Two annotators labeled the remote sensing data and the variance between their sample delineations is used to validate the RandCrowns measure. 

\subsection{Comparisons using multiple annotators}\label{sec:MultipleAnnotators}
We quantitatively and qualitatively compared the RandCrowns and IoU scores based on labels generated using only imagery by multiple annotators. Due to the uncertainty in expert annotations, we do not assume a single set of “true”  ground-truth labels that delineate the tree crowns in our dataset. Instead, we compare the labels created by the four annotators and measure the average variance between each set of three annotators based on treating the fourth as the desired target for comparison. We compute the average variance as,
\begin{equation}
\label{eq:avgvar}
   \Bar{S^2} = \frac{1}{K}\sum_{k=1}^{K}\frac{\sum_{z=1}^{Z}\left(\chi(D_{z,k},T_k)-\mu_{\chi_k}\right)^2}{Z-1},
\end{equation}
where annotations from the three ($Z$) are treated as sample delineations ($D_{z,k}$), while the fourth is the desired target ($T_k$) in each experiment. There are $K$ crowns in an experimental dataset and $\mu_{\chi_k}$ is the average quantitative measurement ($\chi$ representing either IoU, IoUCrowns, or RandCrowns) for the kth crown. As noted in Equation \ref{eq:avgvar}, each individual delineation and assigned target are given a quantitative score. We pair sample delineations to the desired target given the highest IoU score for each delineation-desired target pair, which leads to a conservative assessment of the new method. The main goal in developing the RandCrown metric is to achieve low variance among annotators, but given our target selection method, we also expect the metric to indicate a true delineation. 

Given that the goal is to create a measure which assigns similar annotations values with minimal variance in the presence of uncertainty in labels, we computed the average variance of RandCrowns between annotators and varied the parameters of the RandCrowns measure. With the heuristic we established to compute the true negative outer boundary, there are three parameters to be chosen: the true positive boundary parameter, true negative inner boundary parameter, and the ratio of the true negative area to the true positive area (see Figure \ref{fig:RandCrowns}). We varied these parameters within two experiments. The first compared annotations from four annotators on tree crowns delineated from remote sensing imagery while the second compared annotations from two annotators on polygons derived from field measurements. The parameters selected for both experiments were chosen to: 1) minimize variance between annotators and 2) produce coherent positive and negative regions. 

In addition to evaluating RandCrowns we also evaluated a buffered version of the IoU by disregarding true negatives in the RandCrowns calculations. Using buffers with the IoU instead of the Rand index resulted in higher variance between annotators than when relying on the RandCrowns measure (Table \ref{table:RCvsIoU}). It also resulted in an increased variance even in comparison to using the IoU without buffers due to ignoring true negatives (Table \ref{table:RCvsIoU}). This demonstrates the importance of considering true negatives in this context and so RandCrowns, not the modified IoU (IoUCrowns), was used for all further experiments. 

\begin{table}[t]
\centering
\caption{Variation of RandCrowns and IoU measures across annotators for the cross-validation bounding box experiment.}
\label{table:RCvsIoU}
\begin{tabular}{|c|c|}
\hline
\textbf{Measure}    & \textbf{Variance} \\ \hline
RandCrowns & \textbf{0.008}    \\ \hline
IoU        & 0.022             \\ \hline
IoUCrowns  & 0.040             \\ \hline
\end{tabular}
\end{table}

\begin{figure}[h!]
\subfigure[]{\includegraphics[width=.45\linewidth]{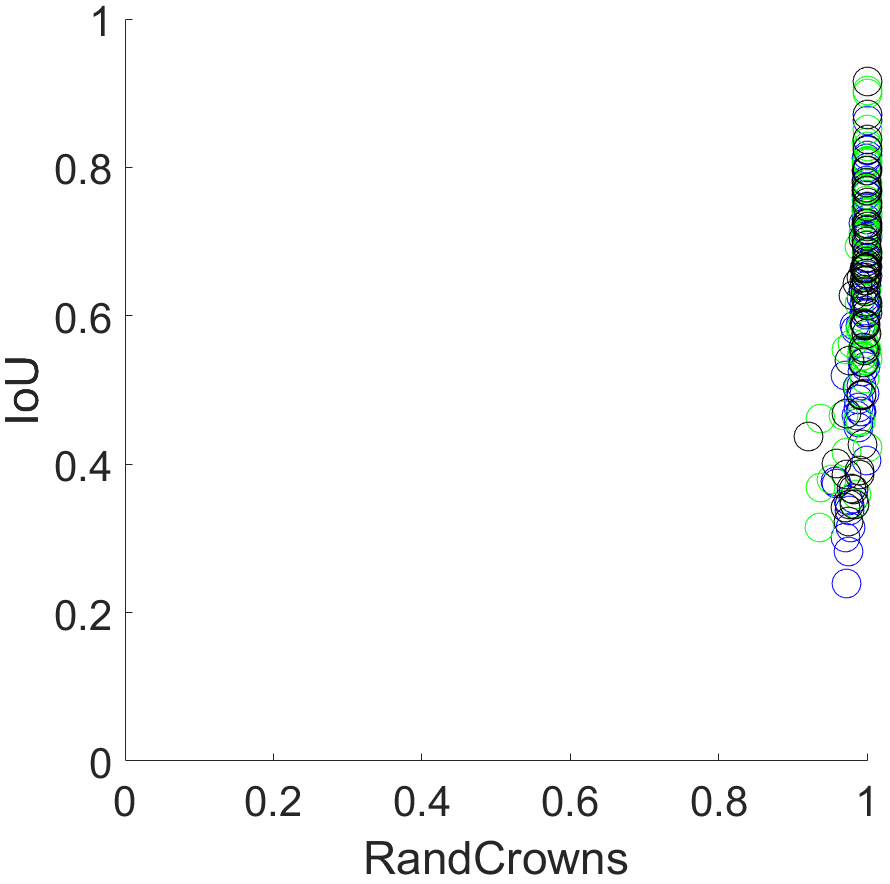} \label{fig:IoUvsRCa}}
\subfigure[]{\includegraphics[width=.45\linewidth]{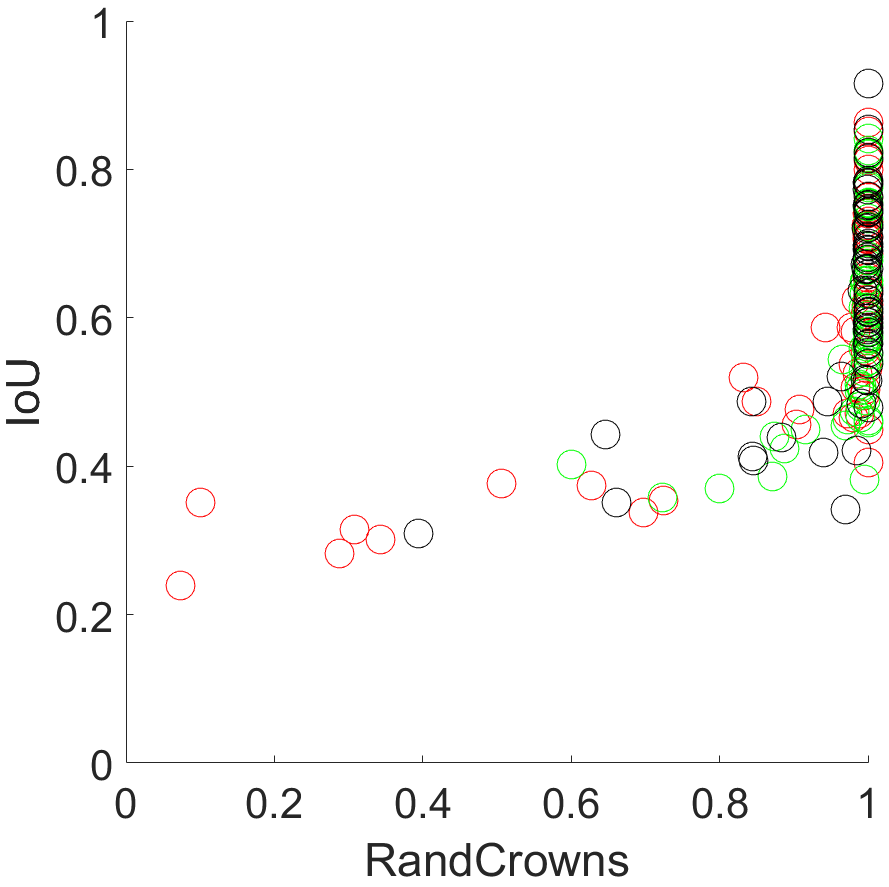} \label{fig:IoUvsRCb}}\\
\subfigure[]{\includegraphics[width=.45\linewidth]{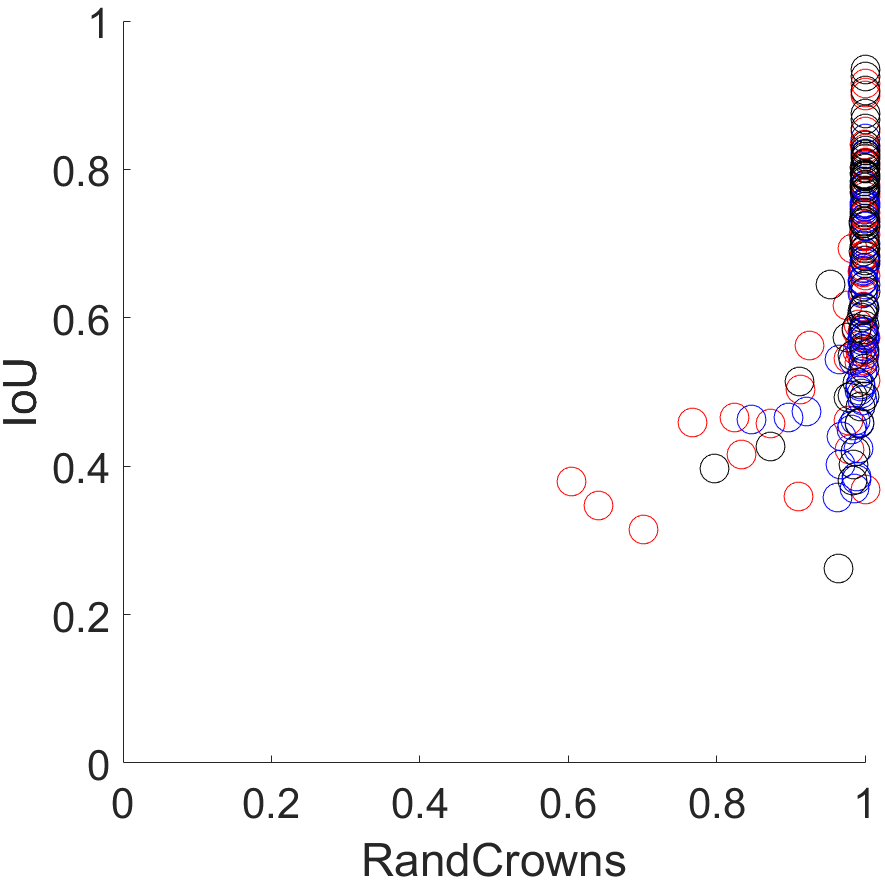} \label{fig:IoUvsRCc}}
\subfigure[]{\includegraphics[width=.45\linewidth]{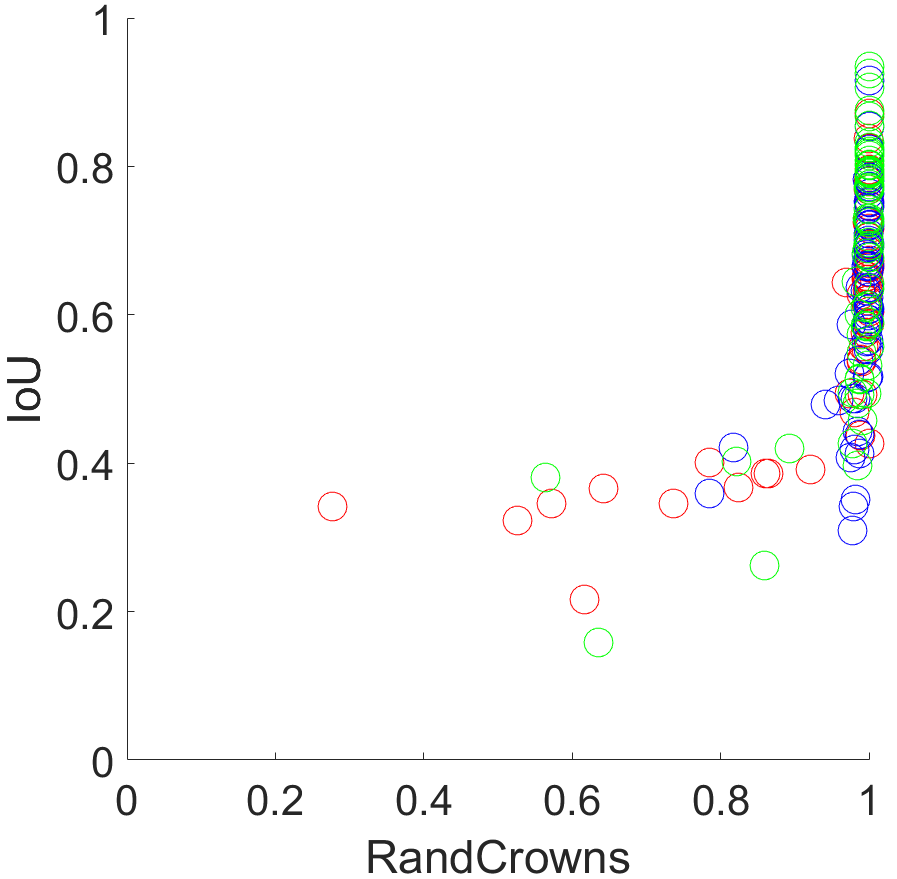} \label{fig:IoUvsRCd}}\\
\begin{center}
\subfigure{\includegraphics[width=.9\linewidth]{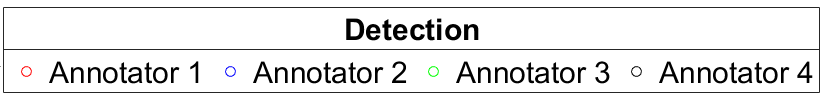}}
\end{center}
\caption{The relationship between the RandCrowns score and the IoU measure for each experiment given our selected parameters. Each series represents an experiment of an annotator-annotator pair where the former is considered the delineation and the latter is the desired target. Each subfigure (a, b, c, and d) corresponds to the annotations from annotators 1-4 being the desired target respectively. Out of the four experiments, only a few examples consist of small RandCrowns scores. Some of these are shown in Figure \ref{fig:RandCrownsLowIoULow}.}
\label{fig:IoUvsRc}
\end{figure}

To investigate the relationship between the IoU and RandCrowns scores we compared each pair of annotator measurements for each cross-validation experiment given our chosen parameters (see Figure \ref{fig:IoUvsRc}). Overall, the RandCrowns is more stable than the IoU with a variance roughly $\frac{1}{3}$ that of the IoU (Table \ref{table:RCvsIoU}). While the RandCrowns assigns matching delineation-desired target pairs with high scores and low variance, the IoU does not. There is no direct relationship between the IoU and RandCrowns for the chosen parameters, but in general RandCrowns scores are higher than IoUs, low RandCrowns scores are associated with low IoU, and most RandCrowns scores are clustered near one.

\begin{figure}
    \centering
    \subfigure{\includegraphics[width=.45\linewidth]{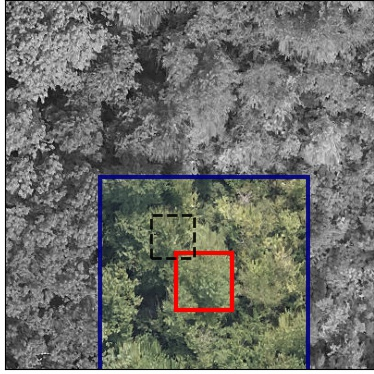}}
    \subfigure{\includegraphics[width=.45\linewidth]{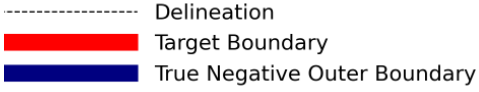}}
    \caption{An example of issues with missed delineations and inappropriate parameters. An example of a missed delineation that has an inappropriately high RC value (RC = 0.99, IoU = 0.02) given parameters that resulted in zero variance between annotators. The zero variance parameters are: true positive boundary 10 cm, true negative inner boundary 30 cm, and area ratio 7. The same delineation with our selected parameters; true positive boundary 70 cm, true negative inner boundary 1.2 m, and area ratio 3 scores a zero as a missed delineation (RC = 0, IoU = 0.02). As the true negative inner boundary parameter is too large with a slightly larger area ratio, all scores are near one even in cases where the delineations clearly do not match the desired target. It is possible that the high RandCrowns measure could be negated by reducing the area of the true positive boundary. However, that would merely fix this type of missed delineation and not necessarily be applicable for other delineations that would encompass small parts of the true positive boundary and much larger areas of an immense false region.
}
    \label{fig:badPars}
\end{figure}

In our experiments we were able to achieve zero variance between annotators using RandCrowns. However, the parameters used to obtain this agreement allowed delineations containing a large false positive region to be given a high score (see Figure \ref{fig:badPars}). Therefore, we used the variance to guide parameter selection and achieved an error between the pair of annotators three times better than the IoU. This highlights the importance of considering uncertainty when comparing multiple annotators for tree crown delineation. Because the IoU only considers an intersection over union of the delineation and the desired target, there is no mechanism to address uncertainty and imprecision.

The parameter ranges for our experiments are shown in Table \ref{table:RCpar}. While there are an infinite number of combinations of parameters to try, we used the average variance between annotators in our cross-validation experiments to guide our choices along with qualitative evaluation. As previously mentioned, some parameter sets may lead to zero variance and we have included an example demonstrating this issue in Figure \ref{fig:badPars}.

\begin{table*}
\centering
\caption{RandCrowns parameters. Each parameter was varied and the average variance across the four cross-validation experiments was recorded. The notation for the parameter ranges are defined to be $\left[l:s:u\right]$, where $l$, $s$, and $u$ correspond to the lower bound, step size, and upper bound respectively. Additionally, visualizations of each selected boundary set were used to qualitatively assess performance.}
\label{table:RCpar}
\begin{tabular}{|l|c|c|c|}
\hline
\multicolumn{4}{|c|}{RandCrowns Parameters}                                                   \\ \hline
Parameter & $\alpha$ & $\omega$ & $\gamma$ \\ \hline
Range     & 10 cm:10 cm:1 m              & 10 cm:10 cm:1.5 m              & 1:7                       \\ \hline
\end{tabular}
\end{table*}

The true positive boundary parameter was varied between 10 cm and 1 m. This reflects an uncertainty of the desired target being 0.01 to 1 m\textsuperscript{2} larger than the actual tree crown. The true negative inner boundary parameter was modified between 10 cm and 1.5 m; which reflects an uncertainty of the crown being 0.01 to 2.25 m\textsuperscript{2} larger than the desired target. Lastly, the ratio of the true negative region to the true positive region ($\gamma$) was varied between 1 and 7. Both quantitatively minimal variance for the cross-validation experiments and qualitative relationships between the selected boundary sizes were used to select the parameters. The lowest average variance with reasonable boundary sizes, 0.008, was observed for true positive boundary size 0.7 m, true negative inner boundary size 1.2 m, and area ratio 3. Although these are the selected parameters, it is possible to obtain a lower variance. If the area ratio is increased solely, the variance can be driven to zero. This case can be visualized in Figure \ref{fig:badPars}.

Given an area ratio that is too large, even with a small true negative inner boundary parameter, the true negatives greatly outnumber the false positives. With the true negative region too large the true negatives greatly outweigh the false positives, resulting in some true positives counted in the delineation even though it is labeling another nearby tree. In order to avoid this issue it is necessary to set reasonable limits on the range of parameters based on the details of the data that address the trade-off between absolute minimal variance and appropriate boundary sizes.

\begin{figure*}
    \centering
    \subfigure[RC = 0.97, IoU = 0.47]{\includegraphics[width=.225\linewidth]{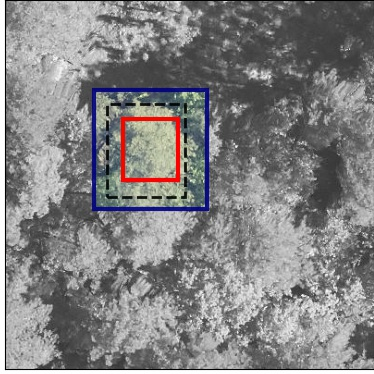}\label{fig:RCHILa}} \hfill
    \subfigure[RC = 1, IoU = 0.24]{\includegraphics[width=.225\linewidth]{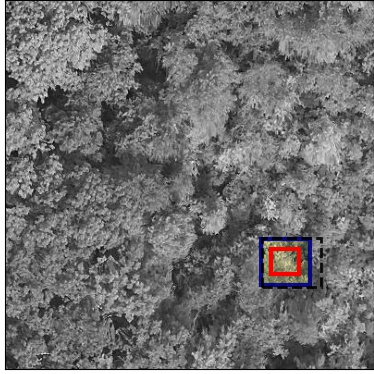}\label{fig:RCHILb}}\hfill
    \subfigure[RC = 0.93, IoU = 0.46]{\includegraphics[width=.225\linewidth]{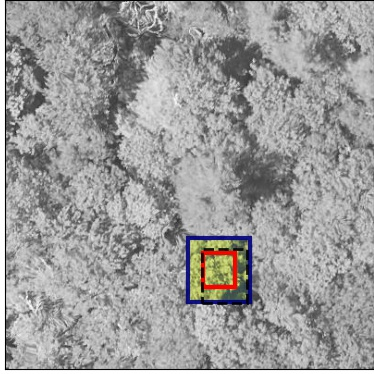}\label{fig:RCHILc}} \hfill
    \subfigure[RC = 0.95 , IoU = 0.46]{\includegraphics[width=.225\linewidth]{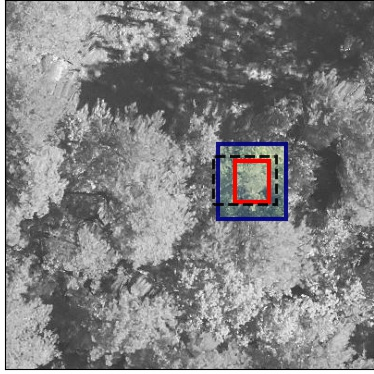}\label{fig:RCHILd}}\\
    \subfigure[RC = 0.97, IoU = 0.24]{\includegraphics[width=.225\linewidth]{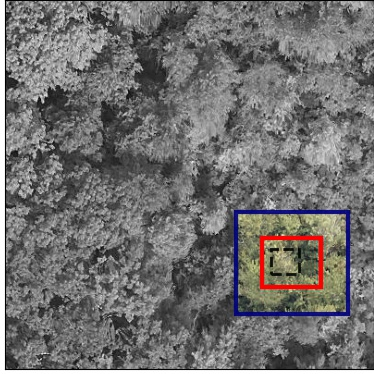}\label{fig:RCHILe}}\hfill
    \subfigure[RC = 0.98, 0.35]{\includegraphics[width=.225\linewidth]{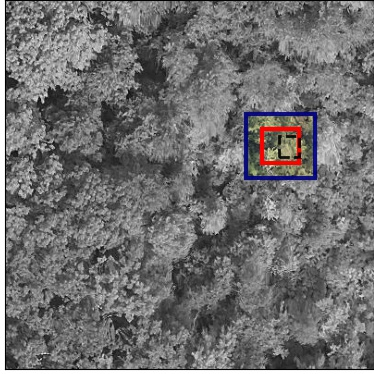}\label{fig:RCHILf}}\hfill
     \subfigure[RC = 0.99, IoU = 0.46]{\includegraphics[width=.225\linewidth]{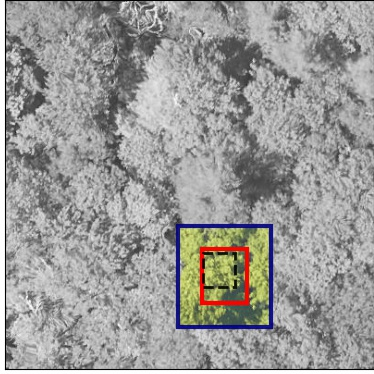}\label{fig:RCHILg}}\hfill
     \subfigure[RC = 0.96, IoU = 0.37]{\includegraphics[width=.225\linewidth]{figures/fig4g.png}\label{fig:RCHILh}}\\
     \begin{center}
    \subfigure{\includegraphics[width=.225\linewidth]{figures/NewLegend_Small.PNG}}
     \end{center}
\caption{Examples where the chosen delineation from the largest IoU value is given a much higher score by the RandCrowns measure. Inner and true negative inner boundaries are not plotted for clarity. Across a variety of desired target sizes, the RandCrowns assigns each of the delineations with a large score. Each case is an example of a delineation that would be considered a false positive given an IoU threshold of 0.5 from  \protect\cite{Lin2017}. Given a smaller threshold of 0.4, half would be considered true positives \protect\cite{Weinstein2020}.}
\label{fig:RandCrownsHighIoULow}
\end{figure*}

For the desired targets from comparing annotators 2 and 4 (Figure \ref{fig:IoUvsRc}b and Figure \ref{fig:IoUvsRc}d) there are multiple examples of delineations which score high with RandCrowns being scored below an IoU score of 0.5 which may be considered false delineations \cite{Lin2017}; some of these interesting cases are shown in Figure \ref{fig:RandCrownsHighIoULow}.

In Figure \ref{fig:RandCrownsHighIoULow}a-d, the delineation is slightly larger than the desired target. While the IoU scores each of these with less than a 0.5 score, the RandCrowns measure assigns high scores greater than 0.9. For each of these example crown delineations, the fact that the desired target (i.e. the “labeled” crown) is larger than the delineation is not penalized, which is desirable because while the area of the crowns differ they both represent the same tree. This supports the importance of the outer region from the true negative inner boundary parameter. Due to the true negative inner boundary parameter being large for smaller crowns, the area between the desired target and the true negative inner boundary is scaled to allow larger delineations to achieve a good score. In Figure \ref{fig:RandCrownsHighIoULow}e-h, the four examples are each cases of smaller delineations than the desired target being granted low scores by the IoU. The IoU scores each of these delineations with less than 0.5 but the RandCrowns measure is greater than 0.96. This shows that although the delineations are not identical to the desired target, they each encompass a significant portion of the true positive boundary while intersecting none of the false delineation area.

\begin{figure}[t]
    \centering
    \subfigure[RC = 1, IoU = 0.9]{\includegraphics[width=.45\linewidth]{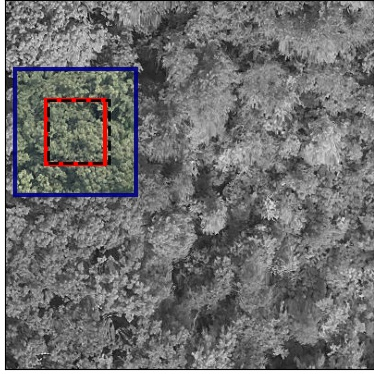}\label{fig:fig5a}} \hfill
    \subfigure[RC = 1, IoU = 0.85]{\includegraphics[width=.45\linewidth]{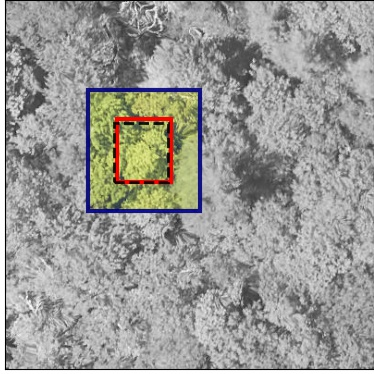}\label{fig:fig5b}}\\
    \subfigure[RC = 1, IoU = 0.93]{\includegraphics[width=.45\linewidth]{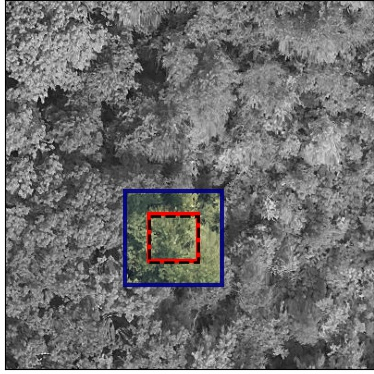}\label{fig:fig5c}} \hfill
    \subfigure[RC = 1, IoU = 0.87]{\includegraphics[width=.45\linewidth]{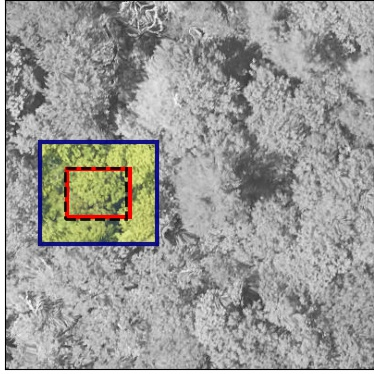}\label{fig:fig5d}} \\
    \begin{center}
        \subfigure{\includegraphics[width=.5\linewidth]{figures/NewLegend_Small.PNG}}
    \end{center}
    \caption{Examples of some of the highest scoring IoU delineations from the cross-validation experiment. In each case the IoU and RandCrowns measure award a high score to the delineation because it is nearly identical to the desired target.}
    \label{fig:RandCrownsHighIoUHigh}
\end{figure}

While in some cases scores differ dramatically between RandCrowns and IoU, in others the IoU and RandCrowns agree on a large score for the delineation. In cases where the IoU is $>0.6$ the RandCrowns score is typically very high, close to one (Figure \ref{fig:IoUvsRc}). Visual inspection of some example crowns in this range demonstrate the difficulty of achieving a high score with IoU unless the delineation is nearly identical to the desired target (Figure \ref{fig:RandCrownsHighIoUHigh}). This means that even small differences among annotators would result in meaningful differences in IoU even with effectively perfect delineation of the crown by an algorithm. Rather than requiring unrealistically perfect annotations, the RandCrowns measure grants a large score to delineations that differ slightly from the desired target but are qualitatively reasonable matches.

\begin{figure}[t]
    \centering
    \subfigure[RC = 0.34, IoU = 0.3]{\includegraphics[width=.45\linewidth]{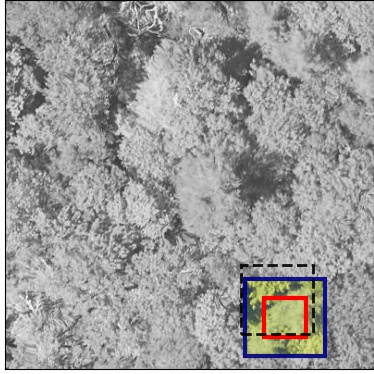}\label{fig:fig6a}} 
    \subfigure[RC = 0.39, IoU = 0.31]{\includegraphics[width=.45\linewidth]{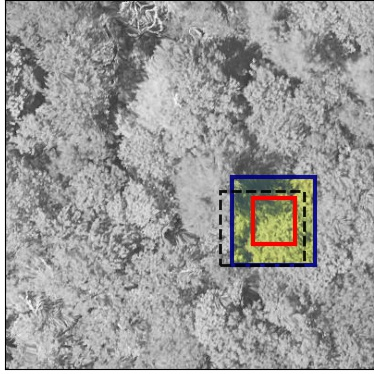}\label{fig:fig6b}}\\
    \subfigure[RC = 0.31, IoU = 0.31]{\includegraphics[width=.45\linewidth]{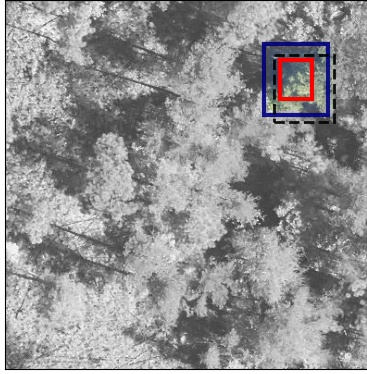}\label{fig:fig6c}} 
    \subfigure[RC = 0.29 , IoU = 0.28]{\includegraphics[width=.45\linewidth]{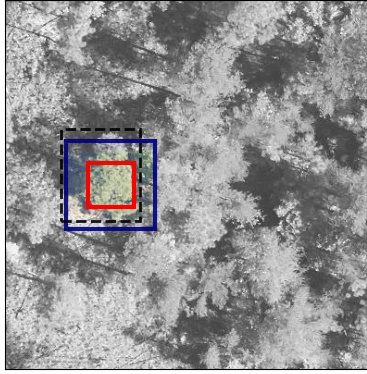}\label{fig:fig6d}} \\
    \begin{center}
        \subfigure{\includegraphics[width=.5\linewidth]{figures/NewLegend_Small.PNG}}
    \end{center}
    \caption{Examples of low scoring delineation-desired target pairs where the RandCrowns and IoU agree on low scores.}
    \label{fig:RandCrownsLowIoULow}
\end{figure}

\begin{figure}
    \centering
    \subfigure[RC = 0.08, IoU = 0.23]{\includegraphics[width=.45\linewidth]{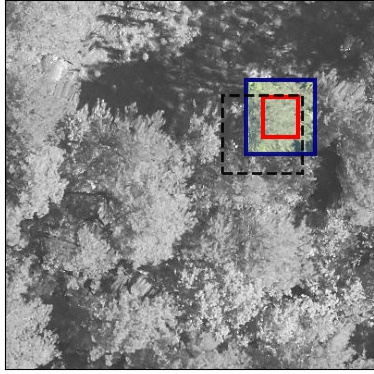}\label{fig:fig7a}} \hfill
    \subfigure[RC = 0.06, IoU = 0.23]{\includegraphics[width=.45\linewidth]{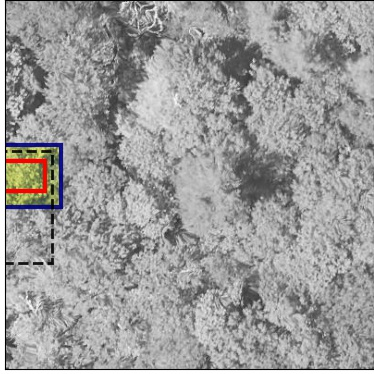}\label{fig:fig7b}}\\
    \subfigure[RC = 0.03, IoU = 0.17]{\includegraphics[width=.45\linewidth]{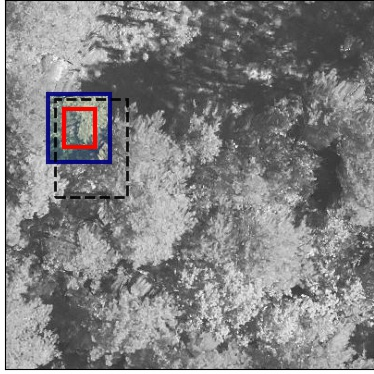}\label{fig:fig7c}} \hfill
    \subfigure[RC = 0, IoU = 0.12]{\includegraphics[width=.45\linewidth]{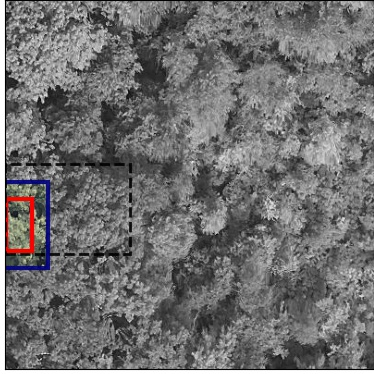}\label{fig:fig7d}} \\
    \begin{center}
        \subfigure{\includegraphics[width=.5\linewidth]{figures/NewLegend_Small.PNG}}
    \end{center}
    \caption{Examples of missed delineations where the RandCrowns and IoU assign low scores. In each case the RandCrowns measure grants a lower score than the IoU.}
    \label{fig:MissedDelineation}
\end{figure}

The RandCrowns measure also scores a variety of poor delineations in agreement with the IoU appropriately (Figure \ref{fig:RandCrownsLowIoULow}). Examples where the IoU is below 0.4 show that the RandCrowns and IoU measures agree with low scores for the four delineation-desired target pairs. Although three of the four delineations shown in Figure \ref{fig:RandCrownsLowIoULow} (a, b, and d) have higher RandCrowns scores, in the case of missed delineations the RandCrowns measure gives lower scores than the IoU as shown in Figure \ref{fig:MissedDelineation}. 

\subsection{Comparing image-only annotations to field-delineated crowns}\label{Sec:ImageToField}
We apply the RandCrowns measure to an experiment comparing multiple annotators against field-delineated crowns. In this context we can consider the field delineated crown to be a perfect delineation (due to the higher accuracy of delineation conducted while in the field; \cite{Graves2018b}) occurring in the presence of uncertainty in the ground-truth labels (represented by the two image-only based delineations by different annotators). Our field dataset includes 22 polygons from 6 plots. The center point of each polygon was displayed to each annotator and the annotator labeled the crown surrounding the centroid with a bounding box to ensure that both annotators were labeling the same tree that had been delineated in the field. We repeat the procedure for the cross-validation experiment and select parameters that minimize the differences in scores between the two annotators. Due to the field-delineated crowns being smaller in this case than the image-based delineation, we found that using smaller true negative inner boundary sizes and area ratios provided less variation between annotators. The average variance between the pair of annotators for RandCrowns was approximately $\frac{1}{14}$  of the variability for IoU (Table \ref{tab:PolygonExperiments}). Our parameters for the polygon implementation were: true positive boundary 60 cm, true negative inner boundary 3 m, and area ratio of 3.

\begin{table}
\centering
\caption{Variation of RandCrowns and IoU measures across annotators for the cross-validation bounding box vs. field-delineated polygons experiment.}
\label{tab:PolygonExperiments}
\begin{tabular}{|c|c|}
\hline
\textbf{Measure}    & \textbf{Variance} \\ \hline
\textbf{RandCrowns} & \textbf{0.001}    \\ \hline
\textbf{IoU}        & 0.014             \\ \hline
\end{tabular}
\end{table}

\begin{figure}
    \centering
    \subfigure[RC = 1, IoU = 0.38]{\includegraphics[width=.45\linewidth,height=3.4cm]{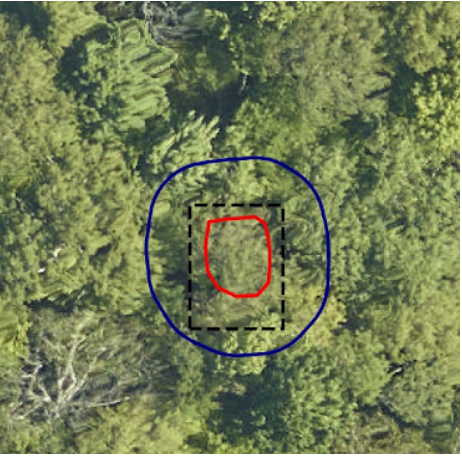}\label{fig:fig8a}} \hfill
    \subfigure[RC = 1, IoU = 0.39]{\includegraphics[width=.45\linewidth,height=3.4cm]{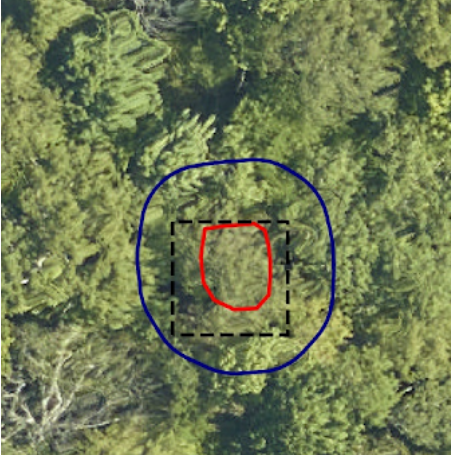}\label{fig:fig8b}}\\
    \subfigure[RC = 1, IoU = 0.48]{\includegraphics[width=.45\linewidth,height=3.4cm]{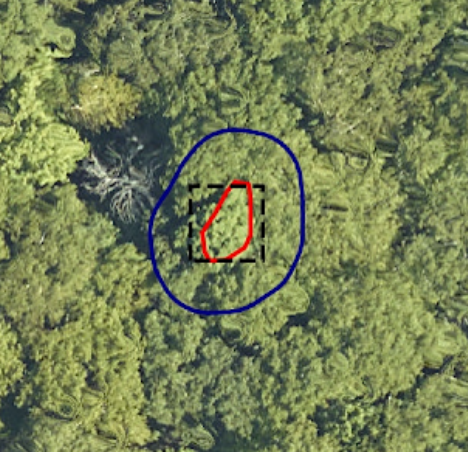}\label{fig:fig8c}} \hfill
    \subfigure[RC = 0.99, IoU = 0.32]{\includegraphics[width=.45\linewidth,height=3.4cm]{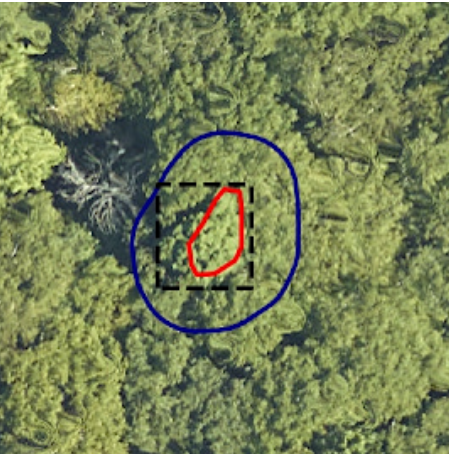}\label{fig:fig8d}} \\
    \begin{center}
        \subfigure{\includegraphics[width=.5\linewidth]{figures/NewLegend_Small.PNG}}
    \end{center}
    \caption{Two comparisons between the annotators where the IoU grants low scores compared the RandCrowns measure. Annotator one labeled (a) and (c) while annotator two labeled (b) and (d). In three of the four cases: (a) (b) and (d), the IoU scores the delineation compared to the desired target with IoU $<$ 0.4. Both crowns were labeled similarly from both bounding box annotators and achieved a zero difference between (a) and (b) and 0.01 between (c) and (d).}
    \label{fig:PolygonExamples}
\end{figure}
Two out of the 22 comparisons ($\sim$10\%) showed low IoU scores ($<$0.5) associated with RandCrowns scores near one (Figure \ref{fig:PolygonExamples}). In both cases the delineation scored poorly based on IoU because the area of the delineation is slightly larger than the desired target. If a threshold of 0.5 \cite{Lin2017} was used to discard missed delineations, all four would be considered false positives. At most, one of these delineations would be considered a true positive given the 0.4 threshold recommended by \cite{Weinstein2020}. In Figure \ref{fig:PolygonExamples}a-b, the scores given by the IoU are almost identical for the two annotators. The same is true of the scores given by the RandCrowns. Meanwhile, in Figure \ref{fig:PolygonExamples}c-d, for nearly equivalent delineations, the IoU has a 0.16 difference in score and the RandCrowns measure is consistent across the annotators. Rather than penalizing labels of the same tree that are not perfect replications of one another, the RandCrowns measure scores the delineations similarly across the annotators. Although the delineations of the two annotators are slightly larger than the desired target, each is labeling the desired crown. Therefore, a large score is warranted in each case and low difference (in this case zero) between the scores is appropriate.

Although we have demonstrated examples of applying the RandCrowns to pairs of delineations, there are cases in application when a global quantitative score is desired. In order to obtain a global RandCrowns score there are many possibilities. To obtain a score for each target, a delineation needs to be assigned to every target. Delineations which do not match any target should reduce the overall global score. Matching a delineation to a target can be done by assigning the target the nearest delineation based on the minimum spatial distance between centers. However, it is possible that multiple delineations have the same distance to a target. In that case, we recommend returning the minimum due to observations from our preliminary results during development of the IDTreeS 2020 competition evaluation toolbox \cite{Stewart2020}. Relying on averaging scores produced by multiple delineations can promote poor delineation performance. Outside of our controlled experiments, if a delineation approach is applied which provides many delineations for each crown, there are multiple options to penalize the delineation method. Either the minimum score can be retained for each target, or the average can be kept and divided by the number of delineations assigned to the target crown. Another possible approach is to score every possible pair of delineation and target pairs and assign each target the candidate delineation with the maximum score. In the latter approach, delineations that are very large may be assigned to multiple crowns and obtain large scores for multiple crowns. We recommend using the first approach for assigning delineations to target crowns. Given a RandCrowns score for each target, the average and standard deviation of the RandCrowns score should be used to compare delineation methods. 

In general, we recommend constructing a validation set consisting of multiple target delineations for each crown. There are several ways of producing multiple target delineations. As we have done in this study, multiple annotators can label each crown individually without knowledge of the other candidate labels. Additionally, target delineations could be proposed by performing warpings (e.g., scalings, rotations, translations) on each hand-labeled delineation. This approach should be investigated in future work. Given a set of target delineations for each crown, the RandCrowns parameters should be varied and those that return minimal average variance of RandCrowns score across the validation set should be proposed as possible parameter settings. 

We demonstrate an application of our new RandCrowns measure to tree crown delineation and provide improvements over the existing IoU. However, several potential limitations should be discussed. Our parameter selection approach does not necessarily provide the optimal parameters for this problem. Certainly, there are other sets of parameter choices which would grant small variance between annotators. While we choose parameters that minimize variance between annotators, qualitative checks are necessary to confirm. As mentioned in the previous paragraph, methods of synthetically generating reliable desired target delineations should be investigated in future work. Additionally, approaches to aggregate the RandCrowns score for a global measure should be tested. Our preliminary work revealed possible issues with assigning a candidate delineation to a target. We provide brief guidance on penalizing methods which produce an inappropriately large number of delineations or delineations of extreme size, however, there are more possibilities that should be studied.

\section{Conclusion}\label{sec:Conclusion}
This paper introduces the RandCrowns measure, an adaptation of the Rand index for weakly-labeled crown delineation. RandCrowns relies on new regions where each term of the Rand index is applied for imprecisely labeled desired targets containing uncertainty. A cross-validation experiment with four annotators was used to demonstrate the ability of the RandCrowns measure to reduce the variance in scores resulting from difference in human annotators with minimal average variance (0.008) in comparison to the commonly used IoU (0.022) and qualitative examples were provided to show the necessity of selecting useful parameters. The RandCrowns measure can also be applied to smoother polygons in addition to bounding boxes, as demonstrated by the second experiment. The parameters were selected to produce minimal variance between the two labelers and the variance was a factor of 14 better than the IoU. In this work we applied RandCrowns for tree crown delineation. However, RandCrowns can be used to quantify segmentation performance in any application that relies on imprecise labels. 

\section*{Acknowledgment}\label{ACKNOWLEDGEMENTS}
Thanks to Ira Harmon and Daisy Wang for helpful discussions of evaluation methods related to tree crown delineation. This research was supported by the Gordon and Betty Moore Foundation’s Data-Driven Discovery Initiative through grant GBMF4563 to E.P. White and by the National Science Foundation through grant 1926542 to E.P. White, S.A. Bohlman, A. Zare, D.Z. Wang, and A. Singh and grant 1442280 to S. A. Bohlman.

\ifCLASSOPTIONcaptionsoff
  \newpage
\fi

\bibliographystyle{IEEEtran}
\begin{spacing}{0.9}
\bibliography{bibliography.bib}
\end{spacing}

\end{document}